\documentclass[12pt,twoside]{article}
\usepackage{latexsym,amsmath,amssymb,hyperref}
\begin{document}
\sloppy\lineskip=0pt
%-------------------------------%
%     My Math-Environments      %
%-------------------------------%
\def\,{\mskip 3mu} \def\>{\mskip 4mu plus 2mu minus 4mu} \def\;{\mskip 5mu plus 5mu} \def\!{\mskip-3mu}
\def\dispmuskip{\thinmuskip= 3mu plus 0mu minus 2mu \medmuskip=  4mu plus 2mu minus 2mu \thickmuskip=5mu plus 5mu minus 2mu}
\def\textmuskip{\thinmuskip= 0mu                    \medmuskip=  1mu plus 1mu minus 1mu \thickmuskip=2mu plus 3mu minus 1mu}
\textmuskip
\def\beq{\dispmuskip\begin{equation}}    \def\eeq{\end{equation}\textmuskip}
\def\beqn{\dispmuskip\begin{displaymath}}\def\eeqn{\end{displaymath}\textmuskip}
\def\bqa{\dispmuskip\begin{eqnarray}}    \def\eqa{\end{eqnarray}\textmuskip}
\def\bqan{\dispmuskip\begin{eqnarray*}}  \def\eqan{\end{eqnarray*}\textmuskip}

%-------------------------------%
%   Macro-Definitions           %
%-------------------------------%
\newenvironment{keywords}{\centerline{\bf\small
Keywords}\begin{quote}\small}{\par\end{quote}\vskip 1ex}
\def\paradot#1{\vspace{1.3ex plus 0.5ex minus 0.5ex}\noindent{\bf\boldmath{#1.}}}
\def\paranodot#1{\vspace{1.3ex plus 0.5ex minus 0.5ex}\noindent{\bf\boldmath{#1}}}
\def\req#1{(\ref{#1})}
\def\toinfty#1{\smash{\stackrel{#1\to\infty}{\longrightarrow}}}
\def\iffs{\Leftrightarrow}
\def\eps{\varepsilon}
\def\epstr{\epsilon}                    % for empty string
\def\nq{\hspace{-1em}}
\def\odt{{\textstyle{1\over 2}}}
\def\odn{{\textstyle{1\over n}}}
\def\SetR{I\!\!R}\def\SetN{I\!\!N}\def\SetQ{I\!\!\!Q}\def\SetB{I\!\!B}\def\SetZ{Z\!\!\!Z}
\def\M{{\cal M}}                        % Set of prob. distributions
\def\S{{\cal S}}                        % Set of prob. distributions
\def\X{{\cal X}}                        % generic set
\def\Y{{\cal Y}}                        % generic set
\def\qmbox#1{{\quad\mbox{#1}\quad}}
\def\qqmbox#1{{\qquad\mbox{#1}\qquad}}
% stacked (in)equalities (nice and simple, just a little bit too condensed horizontally)
\def\equa{\,\smash{\stackrel{\raisebox{0.8ex}{$\scriptstyle+$}}{\smash=}}\,}
\def\leqa{\,\smash{\stackrel{\raisebox{1ex}{$\scriptstyle\!\!\;+$}}{\smash\leq}}\,}
\def\geqa{\,\smash{\stackrel{\raisebox{1ex}{$\scriptstyle+\!\!\;$}}{\smash\geq}}\,}
\def\eqm{\,\smash{\stackrel{\raisebox{0.6ex}{$\scriptstyle\times$}}{\smash=}}\,}
\def\leqm{\,\smash{\stackrel{\raisebox{1ex}{$\scriptstyle\!\!\;\times$}}{\smash\leq}}\,}
\def\geqm{\,\smash{\stackrel{\raisebox{1ex}{$\scriptstyle\times\!\!\;$}}{\smash\geq}}\,}
\def\th{\theta}
\def\e{{\rm e}}                        % natural e
\def\B{\{0,1\}}
\def\E{{\bf E}}                         % Expectation
\def\P{{\rm P}}                         % Probability
\def\l{\ell}
\def\lb{{\log_2}}
\def\v{\boldsymbol}
\def\text#1{\mbox{\scriptsize #1}}
\def\EE{I\!\!E}
\def\o{\omega}
\def\a{\alpha}
\def\Km{K\!m}
\def\thb{{\th'}}
\def\thp{\th'}
\def\Thb{{\Theta'}}
\def\Thp{\Theta'}
\def\ir{\em}                   % important result

%%%%%%%%%%%%%%%%%%%%%%%%%%%%%%%%%%%%%%%%%%%%%%%%%%%%%%%%%%%%%%%%%
%                      T i t l e - P a g e                      %
%%%%%%%%%%%%%%%%%%%%%%%%%%%%%%%%%%%%%%%%%%%%%%%%%%%%%%%%%%%%%%%%%

\title{\vspace{-3ex}\normalsize\sc Technical Report \hfill IDSIA-03-06
\vskip 2mm\bf\Large\hrule height5pt \vskip 6mm
On the Foundations of \\ Universal Sequence Prediction
\vskip 6mm \hrule height2pt}
\author{{\bf Marcus Hutter}\\[3mm]
\normalsize IDSIA, Galleria 2, CH-6928\ Manno-Lugano, Switzerland\\
\normalsize marcus@idsia.ch \hspace{9ex} http://www.idsia.ch/$^{_{_\sim}}\!$marcus }
\date{8 February 2006}
\maketitle

\begin{abstract}\noindent
Solomonoff completed the Bayesian framework by providing a
rigorous, unique, formal, and universal choice for the model class
and the prior. We discuss in breadth how and in which sense
universal (non-i.i.d.)\ sequence prediction solves various
(philosophical) problems of traditional Bayesian sequence
prediction. We show that Solomonoff's model possesses many
desirable properties: Fast convergence and strong bounds, and in
contrast to most classical continuous prior densities has no zero
p(oste)rior problem, i.e.\ can confirm universal hypotheses, is
reparametrization and regrouping invariant, and avoids the
old-evidence and updating problem. It even performs well (actually
better) in non-computable environments.
\vspace{3ex}
\def\contentsname{\centering\normalsize Contents}
{\parskip=-2.5ex\tableofcontents}
\end{abstract}

\begin{keywords}
Sequence prediction, %
Bayes, %
Solomonoff prior, %
Kolmogorov complexity, %
Occam's razor, %
prediction bounds,
model classes, %
philosophical issues, %
symmetry principle, %
confirmation theory, %
reparametrization invariance, %
old-evidence/updating problem, %
(non)computable environments.
\end{keywords}

\newpage
%%%%%%%%%%%%%%%%%%%%%%%%%%%%%%%%%%%%%%%%%%%%%%%%%%%%%%%%%%%%%%%
\section{Introduction}\label{secInt}
%%%%%%%%%%%%%%%%%%%%%%%%%%%%%%%%%%%%%%%%%%%%%%%%%%%%%%%%%%%%%%%

%-------------------------------%
\paradot{Examples and goal}
%-------------------------------%
Given the weather in the past, what is the probability of rain
tomorrow? What is the correct answer in an IQ test asking to
continue the sequence 1,4,9,16,? Given historic stock-charts, can
one predict the quotes of tomorrow? Assuming the sun rose 5000
years every day, how likely is doomsday (that the sun does not
rise) tomorrow? These are instances of the important problem of
inductive inference or time-series forecasting or sequence
prediction. Finding prediction rules for every particular (new)
problem is possible but cumbersome and prone to disagreement or
contradiction. What we are interested in is a formal general
theory for prediction.

%-------------------------------%
\paradot{Bayesian sequence prediction}
%-------------------------------%
The Bayesian framework is the most consistent and successful
framework developed thus far \cite{Earman:93}. A Bayesian
considers a set of environments=\-hypotheses=\-models $\M$ which
includes the true data generating probability distribution $\mu$.
From one's prior belief $w_\nu$ in environment $\nu\in\M$ and the
observed data sequence $x=x_1...x_n$, Bayes' rule yields one's
posterior confidence in $\nu$. In a predictive setting, one
directly determines the predictive probability of the next symbol
$x_{n+1}$ without the intermediate step of identifying a (true or
good or causal or useful) model. Note that classification and
regression can be regarded as special sequence prediction
problems, where the sequence $x_1y_1...x_ny_nx_{n+1}$ of
$(x,y)$-pairs is given and the class label or function value
$y_{n+1}$ shall be predicted.

%-------------------------------%
\paradot{Universal sequence prediction}
%-------------------------------%
The Bayesian framework leaves open how to choose the model class
$\M$ and prior $w_\nu$. General guidelines are that $\M$ should be
small but large enough to contain the true environment $\mu$, and
$w_\nu$ should reflect one's prior (subjective) belief in $\nu$ or
should be non-informative or neutral or objective if no prior
knowledge is available. But these are informal and ambiguous
considerations outside the formal Bayesian framework.
Solomonoff's \cite{Solomonoff:64} rigorous, essentially unique,
formal, and universal solution to this problem is to consider a
single large universal class $\M_U$ suitable for {\em all}
induction problems. The corresponding universal prior $w_\nu^U$ is
biased towards simple environments in such a way that it
dominates=\-superior to all other priors. This leads to an a
priori probability $M(x)$ which is equivalent to the probability
that a universal Turing machine with random input tape outputs
$x$.

%-------------------------------%
\paradot{History and motivation}
%-------------------------------%
Many interesting, important, and deep results have been proven for
Solomonoff's universal distribution $M$
\cite{Zvonkin:70,Solomonoff:78,Li:97,Hutter:04uaibook}.
The motivation and goal of this paper is to provide a broad
discussion of how and in which sense universal sequence prediction
solves all kinds of (philosophical) problems of Bayesian sequence
prediction, and to present some recent results.
Many arguments and ideas could be further developed. I hope that
the exposition stimulates such a future, more detailed, investigation.

%-------------------------------%
\paradot{Contents}
%-------------------------------%
In Section \ref{secBSP} we review the excellent predictive
performance of Bayesian sequence prediction for generic
(non-i.i.d.)\ countable and continuous model classes.
Section \ref{secPrior} critically reviews the classical principles
(indifference, symmetry, minimax) for obtaining objective priors,
introduces the universal prior inspired by Occam's razor
and quantified in terms of Kolmogorov complexity.
In Section \ref{secIID} (for i.i.d.\ $\M$) and Section
\ref{secUSP} (for universal $\M_U$) we show various desirable
properties of the universal prior and class (non-zero p(oste)rior,
confirmation of universal hypotheses, reparametrization and
regrouping invariance, no old-evidence and updating problem) in
contrast to (most) classical continuous prior densities. Finally,
we show that the universal mixture performs better than classical
continuous mixtures, even in uncomputable environments.
Section \ref{secDisc} contains critique and summary.

%%%%%%%%%%%%%%%%%%%%%%%%%%%%%%%%%%%%%%%%%%%%%%%%%%%%%%%%%%%%%%%
\section{Bayesian Sequence Prediction}\label{secBSP}
%%%%%%%%%%%%%%%%%%%%%%%%%%%%%%%%%%%%%%%%%%%%%%%%%%%%%%%%%%%%%%%

%-------------------------------%
\paradot{Notation}
%-------------------------------%
% strings
We use letters $t,n\in\SetN$ for natural numbers, and
denote the cardinality of a set $\cal S$ by $\#{\cal S}$ or $|{\cal S}|$.
We write $\X^*$ for the set of finite strings over some alphabet
$\X$, and $\X^\infty$ for the set of infinite sequences.
For a string $x\in\X^*$ of length $\l(x)=n$ we write
$x_1x_2...x_n$ with $x_t\in\X$, and further abbreviate
$x_{t:n}:=x_t x_{t+1}...x_{n-1}x_n$ and $x_{<n}:=x_1... x_{n-1}$.
%
% probability, expectation, mixtures
We assume that sequence $\o=\o_{1:\infty}\in\X^\infty$ is sampled
from the ``true'' probability measure $\mu$, i.e.\
$\mu(x_{1:n}):=\P[\o_{1:n}=x_{1:n}|\mu]$ is the $\mu$-probability that
$\o$ starts with $x_{1:n}$. We denote expectations
w.r.t.\ $\mu$ by $\E$. In particular for a function $f:\X^n\to\SetR$, we have
$\E[f]=\E[f(\o_{1:n})]=\sum_{x_{1:n}}\mu(x_{1:n})f(x_{1:n})$.
If $\mu$ is unknown but known to belong to a countable class of
environments=models=measures $\M=\{\nu_1,\nu_2,...\}$, and
$\{H_\nu:\nu\in\M\}$ forms a mutually exclusive and complete class
of hypotheses, and $w_\nu:=\P[H_\nu]$ is our prior belief in
$H_\nu$, then $\xi(x_{1:n}):=\P[\o_{1:n}=x_{1:n}] =
\sum_{\nu\in\M}\P[\o_{1:n}=x_{1:n}|H_\nu]\P[H_\nu]$ must be our
(prior) belief in $x_{1:n}$, and
$w_\nu(x_{1:n}):=\P[H_\nu|\o_{1:n}=x_{1:n}]=
{\P[\o_{1:n}=x_{1:n}|H_\nu]\P[H_\nu]\over\P[\o_{1:n}=x_{1:n}]}$ be
our posterior belief in $\nu$ by Bayes' rule.
%
% Asymptotic notation
For a sequence $a_1, a_2, ...$ of random variables,
$\sum_{t=1}^\infty\E[a_t^2]\leq c<\infty$ implies $a_t\toinfty{t} 0$
with $\mu$-probability 1 (w.p.1). Convergence is rapid in the
sense that the probability that $a_t^2$ exceeds $\eps>0$ at more
than ${c\over\eps\delta}$ times $t$ is bounded by $\delta$.
We sometimes loosely call this the number of errors.

%------------------------------%
\paradot{Sequence prediction}
%------------------------------%
Given a sequence $x_1x_2...x_{t-1}$,
we want to predict its likely continuation $x_t$. We assume that
the strings which have to be continued are drawn from a ``true''
probability distribution $\mu$.
The maximal prior information a prediction algorithm can possess
is the exact knowledge of $\mu$, but often the true distribution
is unknown. Instead, prediction is based on a guess $\rho$ of
$\mu$. While we require $\mu$ to be a measure, we allow $\rho$ to
be a semimeasure \cite{Li:97,Hutter:04uaibook}:\footnote{Readers
unfamiliar or uneasy with {\em semi}measures can without loss
ignore this technicality.}
Formally, $\rho:\X^*\to[0,1]$ is a semimeasure if
$\rho(x)\geq\sum_{a\in\X}\rho(xa)\,\forall x\in\X^*$, and a
(probability) measure if equality holds and $\rho(\epstr)=1$,
where $\epstr$ is the empty string. $\rho(x)$ denotes the
$\rho$-probability that a sequence starts with string $x$.
Further, $\rho(a|x):=\rho(xa)/\rho(x)$ is the ``posterior'' or
``predictive'' $\rho$-probability that the next symbol is
$a\in\X$, given sequence $x\in\X^*$.

%------------------------------%
\paradot{Bayes mixture}
%------------------------------%
We may know or assume that $\mu$ belongs to some countable class
${\cal M}:=\{\nu_1,\nu_2,...\}\ni\mu$ of semimeasures. Then we can
use the weighted average on $\cal M$ (Bayes-mixture, data
evidence, marginal)
\beq\label{xidef}
  \xi(x) :=
  \sum_{\nu\in\cal M}w_\nu\!\cdot\!\nu(x),\quad
  \sum_{\nu\in\cal M}w_\nu \leq 1,\quad w_\nu>0.
\eeq
for prediction. The most important property of semimeasure $\xi$
is its dominance
\beq\label{xidom}
  \xi(x) \;\geq\; w_\nu\nu(x) \quad \forall x \;\mbox{and}\; \forall\nu\!\in\!\M,
  \qmbox{in particular} \xi(x) \;\geq\; w_\mu\mu(x)
\eeq
which is a strong form of absolute continuity.

%------------------------------%
\paradot{Convergence for deterministic environments}
%------------------------------%
In the predictive setting we are not interested in identifying the
true environment, but to predict the next symbol well. Let us
consider deterministic $\mu$ first. An environment is called
deterministic if $\mu(\a_{1:n})=1 \forall n$ for some sequence
$\a$, and $\mu=0$ elsewhere (off-sequence). In this case we
identify $\mu$ with $\a$ and the following holds:
\beq\label{detxibnd}\textstyle
  \sum_{t=1}^\infty|1\!-\!\xi(\a_t|\a_{<t})| \;\leq\; \ln w_\a^{-1} \qqmbox{and}
  \xi(\a_{t:n}|\a_t)\to 1 \qmbox{for} n\geq t\to \infty
\eeq
where $w_\a>0$ is the weight of $\a\widehat=\mu\in\M$. This shows
that $\xi(\a_t|\a_{<t})$ rapidly converges to 1 and hence also
$\xi(\bar \a_t|\a_{<t})\to 0$ for $\bar\a_t\neq\a_t$, and that
$\xi$ is also a good multi-step lookahead predictor.
%
%\paradot{Proof}
Proof: $\xi(\a_{1:n})\to c>0$, since $\xi(\a_{1:n})$ is monotone
decreasing in $n$ and $\xi(\a_{1:n})\geq w_\mu\mu(\a_{1:n})=w_\mu>0$.
Hence $\xi(\a_{1:n})/\xi(\a_{1:t})\to c/c=1$ for any limit
sequence $t,n\to\infty$. The bound follows from $\sum_{t=1}^n
1-\xi(x_t|x_{<t})
  \leq - \sum_{t=1}^n\ln \xi(x_t|x_{<t})
  =    -\ln \xi(x_{1:n})$
and $\xi(\a_{1:n})\geq w_\a$.

%------------------------------%
\paradot{Convergence in probabilistic environments}
%------------------------------%
In the general probabilistic case we want to know how close
$\v\xi_t:=\xi(\,\cdot\,_t|\o_{<t})\in\SetR^{|\X|}$ is to the true
probability $\v\mu_t:=\mu(\,\cdot\,_t|\o_{<t})$. One can show that
\beq\label{hbnd}\textstyle
  \sum_{t=1}^n\E[s_t]
  \;\leq\; D_n(\mu||\xi) := \textstyle \E[\ln{\mu(\o_{1:n})\over\xi(\o_{1:n})}]
  \;\leq\; \ln w_\mu^{-1},
\eeq
where $s_t=s_t(\v\mu_t,\v\xi_t)$ can be the squared Euclidian or
Hellinger or absolute or KL distance between $\v\mu_t$ and
$\v\xi_t$, or the squared Bayes-regret
\cite{Hutter:04uaibook}.
The first inequality actually holds for any two (semi)measures,
and the last inequality follows from \req{xidom}.
These bounds (with $n=\infty$) imply
\beqn
 \mbox{$\xi(x_t|\o_{<t})-\mu(x_t|\o_{<t})\to 0$
 for any $x_t$ rapid w.p.1 for $t\to\infty$}.
\eeqn
One can also show multi-step lookahead convergence
$\xi(x_{t:n_t}|\o_{<t})-\mu(x_{t:n_t}|\o_{<t})\to 0$, (even for
unbounded horizon $1\leq n_t-t+1\to\infty$) which is interesting
for delayed sequence prediction and in reactive environments
\cite{Hutter:04uaibook}.

%------------------------------%
\paradot{Continuous environmental classes}
%------------------------------%
The bounds above remain approximately valid for most parametric
model classes.
Let $\M:=\{\nu_\th:\th\in\Theta\subseteq \SetR^d\}$
be a family of probability distributions parameterized by a
$d$-dimensional continuous parameter $\th$, and $\mu \equiv
\nu_{\th_0} \in \M$ the true generating distribution. For a
continuous weight density\footnote{$w()$ will always denote
densities, and $w_{()}$ probabilities.} $w(\th)>0$
the sums \req{xidef} are
naturally replaced by integrals:
\beq\label{xidefc}
  \xi(x) \;:=\; \int_\Theta \!
  w(\th)\!\cdot\!\nu_\th(x)\,d\th, \qquad
  \int_\Theta \! w(\th) \,d\th \;=\; 1 \qquad
\eeq
The most important property of $\xi$ was the dominance \req{xidom}
achieved by dropping the sum over $\nu$. The analogous
construction here is to restrict the integral over $\th$ to a
small vicinity of $\th_0$. Since a continuous parameter can
typically be estimated to accuracy $\propto n^{-1/2}$ after $n$
observations, the largest volume in which $\nu_\th$ as a function of $\th$ is
approximately flat is $\propto (n^{-1/2})^d$, hence
$\xi(x_{1:n})\gtrsim n^{-d/2}w(\th_0)\mu(x_{1:n})$. Under some
weak regularity conditions one can prove
\cite{Clarke:90,Hutter:04uaibook}
\beq\label{cbnd}
  D_n(\mu||\xi) \;:=\;
  \E\textstyle
  \ln{\mu(\o_{1:n}) \over \xi(\o_{1:n})} \;\;\leq\;\;
  \ln w(\th_0)^{-1} + {d\over 2}\ln{n\over 2\pi} +
  {1\over 2}\ln\det\bar\jmath_n(\th_0) + o(1)
\eeq
where $w(\th_0)$ is the weight density (\ref{xidefc}) of $\mu$ in
$\xi$, and $o(1)$ tends to zero for $n \to\infty$, and the average
Fisher information matrix $\bar\jmath_n(\th) = -{1\over
n}\E[\nabla_\th\nabla^T_\th\ln\nu_\th(\o_{1:n})]$ measures the
local smoothness of $\nu_\th$ and is bounded for many reasonable
classes, including all stationary ($k^{th}$-order) finite-state
Markov processes.
We see that in the continuous case, $D_n$ is no longer bounded by
a constant, but grows very slowly (logarithmically) with $n$,
which still implies that $\eps$-deviations are exponentially
seldom. Hence, \req{cbnd} allows to bound \req{hbnd} even in case
of continuous $\M$.

%%%%%%%%%%%%%%%%%%%%%%%%%%%%%%%%%%%%%%%%%%%%%%%%%%%%%%%%%%%%%%%
\section{How to Choose the Prior}\label{secPrior}
%%%%%%%%%%%%%%%%%%%%%%%%%%%%%%%%%%%%%%%%%%%%%%%%%%%%%%%%%%%%%%%

%------------------------------%
\paradot{Classical principles}
%------------------------------%
The probability axioms (implying Bayes' rule) allow to compute
posteriors and predictive distributions from prior ones, but are
mute about how to choose the prior. Much has been written on the
choice of non-informative\-=neutral\-=objective priors (see
\cite{Kass:96} for a survey and references; in Section
\ref{secDisc} we briefly discuss how to incorporate subjective
prior knowledge). For finite $\M$, Laplace's {\em symmetry or
indifference argument} which sets $w_\nu={1\over|\M|}$
$\forall\nu\in\M$ is a reasonable principle. The analogue uniform
density $w(\th)=[\mbox{Vol}(\Theta)]^{-1}$ for a compact
measurable parameter space $\Theta$ is less convincing, since $w$
becomes non-uniform under different parametrization (e.g.\
$\th\leadsto\thp:=\sqrt{\th}$). Jeffreys' solution is to find a
symmetry group of the problem (like permutations for finite $\M$
or translations for $\Theta=\SetR$) and require the prior to be
{\em invariant under group transformations}. Another solution is
the {\em minimax approach} by Bernardo \cite{Clarke:90} which
minimizes (the quite tight) bound \req{cbnd} for the worst
$\mu\in\M$. Choice $w(\th)\propto\sqrt{\det\bar\jmath_n(\th)}$
equalizes and hence minimizes \req{cbnd}. Problems are that there
may be no obvious symmetry, the resulting prior can be improper,
depend on which parameters are treated as nuisance parameters, on
the model class, and on $n$. Other principles are {\em maximum
entropy} and {\em conjugate priors}. The principles above,
although not unproblematic, {\em can} provide good objective
priors in many cases of small discrete or compact spaces, but we
will meet some more problems later. For ``large'' model classes we
are interested in, i.e.\ countably infinite, non-compact, or
non-parametric spaces, the principles typically do not apply or
break down.

%------------------------------%
\paradot{Occam's razor et al}
%------------------------------%
Machine learning, the computer science branch of statistics, often
deals with very large model classes. Naturally, machine learning
has (re)discovered and exploited quite different principles for
choosing priors, appropriate for this situation. The overarching
principles put
together by Solomonoff \cite{Solomonoff:64} are: %
Occam's razor (choose the simplest model consistent with the data), %
Epicurus' principle of multiple explanations (keep all explanations consistent with the data), %
(Universal) Turing machines (to compute, quantify and assign codes to all quantities of interest), and %
Kolmogorov complexity (to define what simplicity/complexity means).

We will first ``derive'' the so called universal prior, and
subsequently justify it by presenting various welcome theoretical
properties and by examples.
The idea is that a priori, i.e.\ before seeing the data, all
models are ``consistent,$\!$'' so a-priori Epicurus would regard all
models (in $\M$) possible, i.e.\ choose $w_\nu>0$
$\forall\nu\in\M$. In order to also do (some) justice to Occam's
razor we should {\em prefer} simple hypothesis, i.e.\ assign high
prior/low prior $w_\nu$ to simple/complex hypotheses $H_\nu$. Before we
can define this prior, we need to quantify the notion of
complexity.

%------------------------------%
\paradot{Notation}
%------------------------------%
% Computability concepts
A function $f:\S\to\SetR\cup\{\pm\infty\}$ is said to be
lower semi-computable (or enumerable) if the set
$\{(x,y)\,:\,y<f(x),\, x\in\S,\, y\in\SetQ\}$ is recursively
enumerable. $f$ is upper semi-computable (or co-enumerable) if
$-f$ is enumerable. $f$ is computable (or recursive)
if $f$ and $-f$ are enumerable. The set of (co)enumerable
functions is recursively enumerable.
% Constants of reasonable size
We write $O(1)$ for a constant of reasonable size: $100$ is
reasonable, maybe even $2^{30}$, but $2^{500}$ is not. We write
$f(x)\leqa g(x)$ for $f(x)\leq g(x)+O(1)$ and $f(x)\leqm g(x)$ for
$f(x)\leq 2^{O(1)}\cdot g(x)$. Corresponding equalities hold if
the inequalities hold in both directions.\footnote{We will ignore
this additive/multiplicative fudge in our discussion till Section
\ref{secDisc}.}
% Most $n$
We say that a property $A(n)\in\{true,false\}$ holds for {\em
most} $n$, if $\#\{t\leq n:A(t)\}/n\toinfty{n} 1$.

%------------------------------%
\paradot{Kolmogorov complexity}
%------------------------------%
We can now quantify the complexity of a string.
Intuitively, a string is simple if it can be described in a few
words, like ``the string of one million ones'', and is complex if
there is no such short description, like for a random object whose
shortest description is specifying it bit by bit. We are
interested in effective descriptions, and hence restrict decoders
to be Turing machines (TMs).
Let us choose some universal (so-called prefix) {\em Turing
machine $U$} with binary input=program tape, $\X$ary output tape, and
bidirectional work tape. We can then define the
{\em prefix Kolmogorov complexity} \cite{Li:97} of
string $x$ as the length $\l$ of the shortest binary program $p$ for
which $U$ outputs $x$:
\beqn
  K(x) \;:=\; \min_p\{\l(p): U(p)=x\}.
\eeqn
For non-string objects $o$ (like numbers and functions) we define
$K(o):=K(\langle o\rangle)$, where $\langle o\rangle\in\X^*$ is
some standard code for $o$. In particular, if $(f_i)_{i=1}^\infty$ is
an enumeration of all (co)enumerable functions, we define
$K(f_i)=K(i)$.

An important property of $K$ is that it is nearly independent of
the choice of $U$. More precisely, if we switch from one universal
TM to another, $K(x)$ changes at most by an additive constant
independent of $x$. For reasonable universal TMs, the compiler
constant is of reasonable size $O(1)$.
A defining property of $K:\X^*\to\SetN$ is that it additively
dominates all co-enumerable functions $f:\X^*\to\SetN$ that
satisfy Kraft's inequality $\sum_x 2^{-f(x)}\;\leq\;1$, i.e.\
$K(x)\leqa f(x)$ for $K(f)=O(1)$. The universal TM provides a
shorter prefix code than any other effective prefix code.
$K$ shares many properties with Shannon's entropy
(information measure) $S$, but $K$ is superior to $S$ in many
respects. To be brief, $K$ is an excellent universal complexity
measure, suitable for quantifying Occam's razor.
We need the following properties of $K$:
\begin{list}{$\bullet$}{\parskip=0ex\parsep=0ex\itemsep=0ex\topsep=1ex} %\partopsep=0ex
\item[$a)$] $K$ is not computable, but only upper semi-computable,
\item[$b)$] the upper bound $K(n)\leqa \lb n+2\lb\log n$, \hfill (\refstepcounter{equation}\theequation\label{Kprop}) %
\item[$c)$] Kraft's inequality $\sum_x 2^{-K(x)}\leq1$, which implies $2^{-K(n)}\leq\odn$ for most $n$, %
\item[$d)$] information non-increase $K(f(x))\leqa K(x)+K(f)$ for recursive $f:\X^*\to\X^*$, %
\item[$e)$] $K(x)\leqa -\lb P(x)+K(P)$ if $P:\X^*\to[0,1]$ is enumerable and $\sum_x P(x)\leq 1$, %
\item[$f)$] $\sum_{x:f(x)=y} 2^{-K(x)}\eqm 2^{-K(y)}$ if $f$ is recursive and $K(f)=O(1)$.
\end{list}
Proofs of $(a)-(e)$ can be found in \cite{Li:97}, and the (easy) proof
of $(f)$ in the extended version of this paper.

%------------------------------%
\paradot{The universal prior}
%------------------------------%
We can now quantify a prior biased towards simple models. First,
we quantify the complexity of an environment $\nu$ or hypothesis
$H_{\nu}$ by its Kolmogorov complexity $K(\nu)$. The universal
prior should be a decreasing function in the model's complexity,
and of course sum to (less than) one. Since $K$ satisfies Kraft's
inequality (\ref{Kprop}$c$), this suggests the following choice:
\beq\label{uprior}
  w_\nu \;=\; w^U_\nu \;:=\; 2^{-K(\nu)}
\eeq
For this choice, the bound \req{hbnd} on $D_n$ reads
\beq\label{DnKCbnd}\textstyle
  \sum_{t=1}^\infty \E[s_t] \;\leq\; D_\infty \;\leq\; K(\mu)\ln 2
\eeq
i.e.\ the number of times, $\xi$ deviates from $\mu$ by more than
$\eps>0$ is bounded by $O(K(\mu))$, i.e.\ is proportional to the
complexity of the environment. Could other choices for $w_\nu$
lead to better bounds? The answer is essentially no
\cite{Hutter:04uaibook}: Consider any other reasonable prior
$w'_\nu$, where reasonable means (lower semi)computable with a
program of size $O(1)$. Then, MDL bound (\ref{Kprop}$e$) with
$P()\leadsto w'_{()}$ and $x\leadsto\langle\mu\rangle$ shows
$K(\mu)\leqa-\lb w'_\mu+K(w'_{()})$, hence $\ln w_\mu'\!^{-1}\geqa
K(\mu)\ln 2$ leads (within an additive constant) to a weaker
bound.
A counting argument also shows that $O(K(\mu))$ errors for most
$\mu$ are unavoidable. So this choice of prior leads to
very good prediction.

Even for continuous classes $\M$, we can assign a (proper) universal prior
(not density) $w_\th^U=2^{-K(\th)}>0$ for computable $\th$, and 0
for uncomputable ones. This effectively reduces $\M$ to a discrete
class $\{\nu_\th\in\M:w_\th^U>0\}$ which is typically dense in $\M$.
We will see that this prior has many advantages over the classical
prior densities.

%%%%%%%%%%%%%%%%%%%%%%%%%%%%%%%%%%%%%%%%%%%%%%%%%%%%%%%%%%%%%%%
\section{Independent Identically Distributed Data}\label{secIID}
%%%%%%%%%%%%%%%%%%%%%%%%%%%%%%%%%%%%%%%%%%%%%%%%%%%%%%%%%%%%%%%

%------------------------------%
\paradot{Laplace's rule for Bernoulli sequences}
%------------------------------%
Let $x=x_1x_2...x_n\in\X^n=\B^n$ be generated by a biased coin
with head=1 probability $\th\in[0,1]$, i.e.\ the likelihood of $x$
under hypothesis $H_\th$ is
$\nu_\th(x)=\P[x|H_\th]=\th^{n_1}(1-\th)^{n_0}$, where
$n_1=x_1+...+x_n=n-n_0$. Bayes assumed a uniform
prior density $w(\th)=1$. The evidence is $\xi(x)=\int_0^1
\nu_\th(x)w(\th)\,d\th={n_1!n_0!\over(n+1)!}$ and the posterior
probability weight density $w(\th|x)=\nu_\th(x)w(\th)/\xi(x)
={(n+1)!\over n_1!n_0!}\th^{n_1}(1-\th)^{n_0}$ of $\th$ after
seeing $x$ is strongly peaked around the frequency estimate
$\hat\th={n_1\over n}$ for large $n$. Laplace
asked for the predictive probability $\xi(1|x)$ of observing
$x_{n+1}=1$ after having seen $x=x_1...x_n$, which is
$\xi(1|x)={\xi(x1)\over \xi(x)}={n_1+1\over n+2}$. (Laplace
believed that the sun had risen for $5\,000$ years = $1\,826\,213$ days
since creation, so he concluded that the probability of doom,
i.e.\ that the sun won't rise tomorrow is ${1\over 1826215}$.)
This looks like a reasonable estimate, since it is close to the
relative frequency, asymptotically consistent, symmetric, even
defined for $n=0$, and not overconfident (never assigns
probability 1).

%------------------------------%
\paradot{The problem of zero prior}
%------------------------------%
But also Laplace's rule is not without problems. The
appropriateness of the uniform prior has been questioned in
Section \ref{secPrior} and will be detailed below. Here we discuss
a version of the zero prior problem. If the prior is zero, then
the posterior is necessarily also zero. The above example seems
unproblematic, since the prior and posterior {\em densities}
$w(\th)$ and $w(\th|x)$ are non-zero. Nevertheless it is
problematic e.g.\ in the context of scientific confirmation theory
\cite{Earman:93}.

Consider the hypothesis $H$ that all balls in some urn, or all
ravens, are black (=1). A natural model is to assume that
balls/ravens are drawn randomly from an infinite population with
fraction $\th$ of black balls/ravens and to assume a uniform prior
over $\th$, i.e.\ just the Bayes-Laplace model. Now we draw $n$
objects and observe that they are all black.

We may formalize $H$ as the hypothesis $H':=\{\th=1\}$. Although
the posterior probability of the relaxed hypothesis
$H_\eps:=\{\th\geq 1-\eps\}$, $\P[H_\eps|1^n]=\int_{1-\eps}^1
w(\th|1^n)\,d\th=\int_{1-\eps}^1(n+1)\th^n d\th=1-(1-\eps)^{n+1}$
tends to 1 for $n\to\infty$ for every fixed $\eps>0$,
$\P[H'|1^n]=\P[H_0|1^n]$ remains identically zero, i.e.\ no amount
of evidence can confirm $H'$. The reason is simply that zero prior
$\P[H']=0$ implies zero posterior.

Note that $H'$ refers to the unobservable quantity $\th$ and only
demands blackness with probability 1. So maybe a better
formalization of $H$ is purely in terms of observational
quantities: $H'':=\{\o_{1:\infty}=1^\infty\}$. Since
$\xi(1^n)={1\over n+1}$, the predictive probability of observing
$k$ further black objects is
$\xi(1^k|1^n)={\smash{\xi(1^{n+k})\over\xi(1^n)}}={n+1\over n+k+1}$. While
for fixed $k$ this tends to 1,
$\P[H''|1^n]=\lim_{k\to\infty}\xi(1^k|1^n)\equiv 0$ $\forall n$,
as for $H'$.

One may speculate that the crux is the infinite population. But
for a finite population of size $N$ and sampling with (similarly
without) repetition, $\P[H''|1^n]=\xi(1^{N-n}|1^n)={n+1\over N+1}$
is close to one only if a large fraction of objects has been
observed. This contradicts scientific practice: Although only a
tiny fraction of all existing ravens have been observed, we regard
this as sufficient evidence for believing strongly in $H$.

There are two solutions of this problem: We may abandon
strict\-/logical\-/all-quantified\-/universal hypotheses
altogether in favor of soft hypotheses like $H_\eps$. Although not
unreasonable, this approach is unattractive for several reasons.
The other solution is to assign a non-zero prior to $\th=1$.
Consider, for instance, the improper density
$w(\th)=\odt[1+\delta(1-\th)]$, where $\delta$ is the Dirac-delta
($\int f(\th)\delta(\th-a)\,d\th=f(a)$), or equivalently
$\P[\th\geq a]=1-\odt a$. We get
$\xi(x_{1:n})=\odt[{n_1!n_0!\over(n+1)!}+\delta_{0n_0}]$, where
$\delta_{ij}=\{{1 \text{ if } i=j\atop 0 \text{ else}\;\;\; }\}$
is Kronecker's $\delta$. In particular $\xi(1^n)=\odt{n+2\over
n+1}$ is much larger than for uniform prior. Since
$\xi(1^k|1^n)={n+k+2\over n+k+1}\cdot{n+1\over n+2}$, we get
$\P[H''|1^n]=\lim_{k\to\infty}\xi(1^k|1^n)={n+1\over n+2}\to 1$,
i.e.\ $H''$ gets strongly confirmed by observing a reasonable
number of black objects. This correct asymptotics also follows
from the general result \req{detxibnd}. Confirmation of $H''$ is
also reflected in the fact that $\xi(0|1^n)={1\over(n+2)^2}$ tends
much faster to zero than for uniform prior, i.e.\ the confidence that
the next object is black is much higher. The power actually
depends on the shape of $w(\th)$ around $\th=1$. Similarly $H'$
gets confirmed: $\P[H'|1^n]=\mu_1(1^n)\P[\th=1]/\xi(1^n)={n+1\over
n+2}\to 1$.
On the other hand, if a single (or more) 0 are observed ($n_0>0$),
then the predictive distribution $\xi(\cdot|x)$ and posterior
$w(\th|x)$ are the same as for uniform prior.

The findings above remain qualitatively valid for i.i.d.\
processes over finite non-binary alphabet $|\X|>2$ and for
non-uniform prior.

Surely to get a generally working setup, we should also assign a
non-zero prior to $\th=0$ and to all other ``special'' $\th$, like
$\odt$ and ${1\over 6}$, which may naturally appear in a
hypothesis, like ``is the coin or die fair''. The natural
continuation of this thought is to assign non-zero prior to all
computable $\th$. This is another motivation for the universal
prior $w_\th^U=2^{-K(\th)}$ \req{uprior} constructed in Section
\ref{secPrior}. It is difficult but not impossible to operate with
such a prior \cite{Hutter:04mdlspeed}.
One may want to mix the discrete prior $w_\nu^U$ with a continuous
(e.g.\ uniform) prior density, so that the set of non-computable
$\th$ keeps a non-zero density. Although possible, we will see
that this is actually not necessary.

%------------------------------%
\paradot{Reparametrization invariance}
%------------------------------%
Naively, the uniform prior is justified by the indifference
principle, but as discussed in Section \ref{secPrior}, uniformity
is not reparametrization invariant. For instance if in our
Bernoulli example we introduce a new parametrization
$\thp=\sqrt\th$, then the $\thp$-density $w'(\thp)=2\sqrt{\th}w(\th)$
is no longer uniform if $w(\th)=1$ is uniform.

More generally, assume we have some principle which leads to some
prior $w(\th)$. Now we apply the principle to a different
parametrization $\thp\in\Thp$ and get prior $w'(\th')$. Assume
that $\th$ and $\thp$ are related via bijection
$\th=f(\thp)$. Another way to get a $\thp$-prior is to
transform the $\th$-prior $w(\th)\leadsto \tilde w(\thp)$.
The reparametrization invariance principle (RIP) states that
$w'$ should be equal to $\tilde w$.

For discrete $\Theta$, simply $\tilde w_\thb=w_{f(\thp)}$, and a
uniform prior remains uniform ($w'_\thb=\tilde
w_\thb=w_\th={1\over|\Theta|}$) in any parametrization, i.e.\ the
indifference principle satisfies RIP in finite model classes.

In case of densities, we have $\tilde
w(\thp)=w(f(\thp)){df(\thp)\over d\thp}$, and the indifference
principle violates RIP for non-linear transformations $f$. But
Jeffrey's and Bernardo's principle satisfy RIP. For instance, in
the Bernoulli case we have $\bar\jmath_n(\th)={1\over\th}+{1\over
1-\th}$, hence $w(\th)={1\over\pi}[\th(1-\th)]^{-1/2}$ and
$w'(\thp)={1\over\pi}[f(\thp)(1-f(\thp))]^{-1/2}{df(\thp)\over
d\thp}=\tilde w(\thp)$.

Does the universal prior $w_\th^U=2^{-K(\th)}$ satisfy RIP? If we
apply the ``universality principle'' to a $\thp$-parametrization,
we get $w_\thb'\!\!^U=2^{-K(\thp)}$. On the other hand, $w_\th$
simply transforms to $\tilde w_\thb^U = w_{f(\thp)}^U =
2^{-K(f(\thp))}$ ($w_\th$ is a discrete (non-density) prior, which
is non-zero on a discrete subset of $\M$). For computable $f$ we
have $K(f(\thp))\leqa K(\thp)+K(f)$ by (\ref{Kprop}$d$), and
similarly $K(f^{-1}(\th))\leqa K(\th)+K(f)$ if $f$ is invertible.
Hence for simple bijections $f$ i.e.\ for $K(f)=O(1)$, we have
$K(f(\thp))\equa K(\thp)$, which implies $w_\thb'\!\!^U \eqm
\tilde w_\thb^U$, i.e.\ {\ir the universal prior satisfies RIP}
w.r.t.\ simple transformations $f$ (within a multiplicative
constant).

%------------------------------%
\paradot{Regrouping invariance}
%------------------------------%
There are important transformations $f$ which are {\em not}
bijections, which we consider in the following. A simple
non-bijection is $\th=f(\thp)=\thp^2$ if we consider
$\thp\in[-1,1]$. More interesting is the following example: Assume
we had decided not to record blackness versus non-blackness of
objects, but their ``color''. For simplicity of exposition assume
we record only whether an object is black or white or colored, i.e.\
$\X'=\{B,W,C\}$. In analogy to the binary case we use the
indifference principle to assign a uniform prior on
$\v\thp\in\Thp:=\Delta_3$, where
$\Delta_d:=\{\v\thp\in[0,1]^d:\sum_{i=1}^d\thp_i=1\}$, and
$\nu_\thb(x'_{1:n})=\prod_i{\thp_i}^{n_i}$. All inferences regarding
blackness (predictive and posterior) are identical to the binomial
model $\nu_\th(x_{1:n})=\th^{n_1}(1-\th)^{n_0}$ with $x'_t=B$
$\leadsto$ $x_t=1$ and $x'_t=W\,$or$\,C$ $\leadsto$ $x_t=0$ and
$\th=f(\v\thp)=\thp_B$ and
$w(\th)=\int_{\Delta_3}w'(\v\thp)\delta(\thp_B-\th)d\v\thp$.
Unfortunately, for uniform prior $w'(\thp)\propto 1$, $w(\th)\propto
1-\th$ is {\em not} uniform, i.e.\ the indifference principle is
{\em not} invariant under splitting/grouping, or general
regrouping. Regrouping invariance is regarded as a very important
and desirable property \cite{Walley:96}.

We now consider general i.i.d.\ processes
$\nu_\th(x)=\prod_{i=1}^d\th_i^{n_i}$. Dirichlet priors
$w(\th)\propto\prod_{i=1}^d\th_i^{\a_i-1}$ form a natural
conjugate class ($w(\th|x)\propto\prod_{i=1}^d\th_i^{n_i+\a_i-1}$)
and are the default priors for multinomial (i.i.d.)\ processes over
finite alphabet $\X$ of size $d$. Note that $\xi(a|x)={n_a+\a_a\over
n+\a_1+...+\a_d}$ generalizes Laplace's rule and coincides with
Carnap's \cite{Earman:93} confirmation function.
Symmetry demands $\a_1=...=\a_d$; for instance $\a\equiv 1$ for
uniform and $\a\equiv\odt$ for Bernard-Jeffrey's prior. Grouping
two ``colors'' $i$ and $j$ results in a Dirichlet prior with
$\a_{i\&j}=\a_i+\a_j$ for the group. The only way to respect
symmetry under all possible groupings is to set $\a\equiv 0$. This
is Haldane's improper prior, which results in unacceptably
overconfident predictions $\xi(1|1^n)=1$. Walley \cite{Walley:96}
solves the problem that there is no single acceptable prior
density by considering sets of priors.

We now show that the universal prior $w_\th^U=2^{-K(\th)}$ is
invariant under regrouping, and more generally under all simple
(computable with complexity O(1)) even non-bijective
transformations. Consider prior $w'_\thb$. If $\th=f(\thp)$ then
$w'_\thb$ transforms to $\tilde
w_\th=\sum_{\thp:f(\thp)=\th}w'_\thb$ (note that for
non-bijections there is more than one $w'_\thb$ consistent with
$\tilde w_\th$). In $\thp$-parametrization, the universal prior
reads $w_\thb'\!\!^U=2^{-K(\thp)}$.
Using (\ref{Kprop}$f$) with $x=\langle\thp\rangle$ and
$y=\langle\th\rangle$ we get
$\tilde w_\th^U = \sum_{\thp:f(\thp)=\th}2^{-K(\thp)}
  \eqm 2^{-K(\th)} = w_\th^U$,
i.e.\ {\ir the universal prior is general transformation and hence
regrouping invariant} (within a multiplicative constant) w.r.t.\
simple computable transformations $f$.

Note that reparametrization and regrouping invariance hold
for arbitrary classes $\M$ and are not limited to the i.i.d.\ case.

%%%%%%%%%%%%%%%%%%%%%%%%%%%%%%%%%%%%%%%%%%%%%%%%%%%%%%%%%%%%%%%
\section{Universal Sequence Prediction}\label{secUSP}
%%%%%%%%%%%%%%%%%%%%%%%%%%%%%%%%%%%%%%%%%%%%%%%%%%%%%%%%%%%%%%%

%------------------------------%
\paradot{Universal choice of $\cal M$}
%------------------------------%
The bounds of Section \ref{secBSP} apply if $\M$ contains the true
environment $\mu$. The larger $\M$ the less restrictive is this
assumption. The class of all computable distributions, although
only countable, is pretty large from a practical point of view,
since it includes for instance all of today's valid physics theories.
It is the largest class, relevant from a computational point of
view. Solomonoff \cite[Eq.(13)]{Solomonoff:64} defined and studied
the mixture over this class.

One problem is that this class is not enumerable, since
the class of computable functions $f:\X^*\to\SetR$ is not enumerable
(halting problem), nor is it decidable whether a function is a
measure. Hence $\xi$ is completely incomputable. Levin
\cite{Zvonkin:70} had the idea to ``slightly'' extend the class
and include also lower semi-computable semimeasures. One can show
that this class $\M_U=\{\nu_1,\nu_2,...\}$ is enumerable, hence
\beq\label{xiUdef}
  \xi_U(x) \;=\; \sum_{\nu\in\M_U} w_\nu^U \nu(x)
\eeq
is itself lower semi-computable, i.e.\ $\xi_U\in\M_U$, which is a
convenient property in itself. Note that since ${1\over n\log^2
n}\leqm w_{\nu_n}^U\leq\odn$ for most $n$ by (\ref{Kprop}$b$) and
(\ref{Kprop}$c$), most $\nu_n$ have prior approximately reciprocal
to their index $n$.

In some sense $\M_U$ is the largest class of environments for
which $\xi$ is in some sense computable \cite{Hutter:04uaibook},
but see \cite{Schmidhuber:02gtm} for even larger classes.

%------------------------------%
\paradot{The problem of old evidence}
%------------------------------%
An important problem in Bayesian inference in general and
(Bayesian) confirmation theory \cite{Earman:93} in particular is
how to deal with `old evidence' or equivalently with `new
theories'. How shall a Bayesian treat the case when some evidence
$E\widehat=x$ (e.g.\ Mercury's perihelion advance) is known
well-before the correct hypothesis/theory/model $H\widehat=\mu$ (Einstein's
general relativity theory) is found? How shall $H$ be added to the
Bayesian machinery a posteriori? What is the prior of $H$? Should
it be the belief in $H$ in a hypothetical counterfactual world in
which $E$ is not known? Can old evidence $E$ confirm $H$? After
all, $H$ could simply be constructed/biased/fitted towards
``explaining'' $E$.

The universal class $\M_U$ and universal prior $w_\nu^U$ formally
solve this problem: The universal prior of $H$ is $2^{-K(H)}$.
This is independent of $\M$ and of whether $E$ is known or not. If
we use $E$ to construct $H$ or fit $H$ to explain $E$, this will
lead to a theory which is more complex ($K(H)\geqa K(E)$) than a
theory from scratch ($K(H)=O(1)$), so cheats are automatically
penalized. There is no problem of adding hypotheses to $\M$ a
posteriori. Priors of old hypotheses are not affected. Finally,
$\M_U$ includes {\em all} hypothesis (including yet unknown or
unnamed ones) a priori. So at least theoretically, updating $\M$
is unnecessary.

%------------------------------%
\paradot{Other representations of $\xi_U$}
%------------------------------%
There is a much more elegant representation of $\xi_U$:
Solomonoff \cite[Eq.(7)]{Solomonoff:64} defined the {\em
universal prior} $M(x)$ as the probability that the output
of a universal Turing machine $U$ starts with $x$ when provided
with fair coin flips on the input tape.
Note that a uniform distribution is also used in the so-called
No-Free-Lunch theorems to prove the impossibility of universal
learners, but in our case the uniform distribution is piped
through a universal Turing machine which defeats these negative
implications.
Formally, $M$ can be defined as
\beq\label{Mdef}
  M(x)\;:=\;\sum_{p\;:\;U(p)=x*}\nq 2^{-\l(p)} \;\eqm\; \xi_U(x)
\eeq
where the sum is over all (so-called minimal) programs $p$ for
which $U$ outputs a string starting with $x$.
$M$ may be regarded as a $2^{-\l(p)}$-weighted mixture over all
computable deterministic environments $\nu_p$ ($\nu_p(x)=1$ if $U(p)=x*$ and
0 else).
Now, as a positive surprise, {\ir $M(x)$ coincides with
$\xi_U(x)$} within an irrelevant multiplicative constant. So it is
actually sufficient to consider the class of {\em deterministic}
semimeasures. The reason is that the probabilistic semimeasures
are in the convex hull of the deterministic ones, and so need not
be taken extra into account in the mixture.

%------------------------------%
\paradot{Bounds for computable environments}
%------------------------------%
The bound \req{DnKCbnd} surely is applicable for $\xi=\xi_U$ and
now holds for {\em any} computable measure $\mu$. Within an
additive constant the bound is also valid for $M\eqm\xi$. That is,
{\ir $\xi_U$ and $M$ are excellent predictors with the only
condition that the sequence is drawn from any computable
probability distribution}. Bound \req{DnKCbnd} shows that the
total number of prediction errors is small. Similarly to
\req{detxibnd} one can show that $\sum_{t=1}^n|1-M(x_t|x_{<t})|
\leq \Km(x_{1:n})\ln 2$, where the monotone complexity
$\Km(x):=\min\{\l(p):U(p)=x*\}$ is defined as the length of the
shortest (nonhalting) program computing a string starting with $x$
\cite{Zvonkin:70,Li:97,Hutter:04uaibook}.

If $x_{1:\infty}$ is a computable sequence, then
$\Km(x_{1:\infty})$ is finite, which implies {\ir $M(x_t|x_{<t})\to 1$
on every computable sequence}. This means that if the environment is
a computable sequence (whichsoever, e.g.\ $1^\infty$ or the digits
of $\pi$ or $\e$), after having seen the first few digits, $M$
correctly predicts the next digit with high probability, i.e.\ it
recognizes the structure of the sequence. In particular, observing
an increasing number of black balls or black ravens or sunrises,
$M(1|1^n)\to 1$ ($\Km(1^\infty)=O(1)$) becomes rapidly confident
that future balls and ravens are black and that the sun will rise
tomorrow.

%------------------------------%
\paradot{Universal is better than continuous $\M$}
%------------------------------%
Although we argued that incomputable environments $\mu$ can safely
be ignored, one may be nevertheless uneasy using Solomonoff's
$M\eqm\xi_U$ (\ref{Mdef}) if outperformed by a continuous mixture
$\xi$ (\ref{xidefc}) on such $\mu\in\M\setminus\M_U$, for instance
if $M$ would fail to predict a Bernoulli($\th$) sequence for
incomputable $\th$. Luckily this is not the case: Although
$\nu_\th()$ and $w_\th$ can be incomputable, the studied classes
$\M$ themselves, i.e.\ the two-argument function $\nu_{()}()$, and
the weight function $w_{()}$, and hence $\xi()$, are typically
computable (the integral can be approximated to arbitrary
precision). Hence $M(x)\eqm\xi_U(x)\geq 2^{-K(\xi)}\xi(x)$ by \req{xiUdef}
and $K(\xi)$ is often quite small. This implies for
{\em all} $\mu$
\beqn\textstyle
  D_n(\mu||M)
    \;\equiv\; \E[\ln\!{\mu(\o_{1:n})\over M(\o_{1:n})}]
    \;=\; \E[\ln\!{\mu(\o_{1:n})\over\xi(\o_{1:n})}]
    \!+\! \E[\ln\!{\xi(\o_{1:n})\over M(\o_{1:n})}]
    \;\leqa\; D_n(\mu||\xi) \!+\! K(\xi)\ln 2
\eeqn
So any bound \req{cbnd} for $D_n(\mu||\xi)$ is directly valid also
for $D_n(\mu||M)$, save an additive constant. That is, $M$ is
superior (or equal) to all computable mixture predictors $\xi$
based on any (continuous or discrete) model class $\M$ and weight
$w(\th)$, even if environment $\mu$ is {\em not} computable.
Furthermore, while for essentially all parametric classes,
$D_n(\mu||\xi)\sim{d\over 2}\ln n$ grows logarithmically in $n$ for all
(incl.\ computable) $\mu\in\M$, $D_n(\mu||M)\leq K(\mu)\ln 2$ is
finite for computable $\mu$.
Bernardo's prior even implies a bound for $M$ that is uniform
(minimax) in $\th\in\Theta$. Many other priors based on reasonable
principles (see Section \ref{secPrior} and \cite{Kass:96}) and
many other computable probabilistic predictors $\rho$ are argued
for. The above actually shows that $M$ is superior to all of them.

%%%%%%%%%%%%%%%%%%%%%%%%%%%%%%%%%%%%%%%%%%%%%%%%%%%%%%%%%%%%%%%
\section{Discussion}\label{secDisc}
%%%%%%%%%%%%%%%%%%%%%%%%%%%%%%%%%%%%%%%%%%%%%%%%%%%%%%%%%%%%%%%

%-------------------------------%
\paradot{Critique and problems}
%-------------------------------%
% Prior Knowledge and subjective priors
In practice we often have extra information about the problem at
hand, which could and should be used to guide the forecasting.
One way is to explicate all our prior knowledge $y$ and place it
on an extra input tape of our universal Turing machine $U$, which
leads to the conditional complexity $K(\cdot|y)$. We now assign
``subjective'' prior $w_{\nu|y}^U=2^{-K(\nu|y)}$ to environment
$\nu$, which is large for those $\nu$ that are simple (have short
description) relative to our background knowledge $y$. Since
$K(\mu|y)\leqa K(\mu)$, extra knowledge never misguides (see
\req{DnKCbnd}). Alternatively we could
prefix our observation sequence $x$ by $y$ and use $M(yx)$ for
prediction \cite{Hutter:04uaibook}.

% On constants and $U$ dependence}
Another critique concerns the dependence of $K$ and $M$ on $U$.
Predictions for short sequences $x$ (shorter than typical compiler
lengths) can be arbitrary. But taking into account our
(whole) scientific prior knowledge $y$, and predicting the now long string
$yx$ leads to good (less sensitive to ``reasonable'' $U$)
predictions \cite{Hutter:04uaibook}.

% Incomputability
Finally, $K$ and $M$ can serve as ``gold standards'' which
practitioners should aim at, but since they are only
semi-computable, they have to be (crudely) approximated in
practice. Levin complexity \cite{Li:97}, Schmidhuber's speed
prior, the minimal message and description length principles
\cite{Wallace:05}, and off-the-shelf compressors like Lempel-Ziv
are such approximations, which have been successfully applied to a
plethora of problems \cite{Cilibrasi:05,Schmidhuber:04oops}.

%-------------------------------%
\paradot{Summary}
%-------------------------------%
We compared traditional Bayesian sequence prediction based on
continuous classes and prior densities to Solomonoff's universal
predictor $M$, prior $w_\nu^U$, and class $\M_U$. We discussed:
% pros
Convergence for generic class and prior, %
the relative entropy bound for continuous classes, %
indifference/symmetry principles, %
the problem of zero p(oste)rior and confirmation of universal hypotheses, %
reparametrization and regrouping invariance, %
the problem of old evidence and updating, %
that $M$ works even in non-computable environments, %
% cons
how to incorporate prior knowledge, %
the prediction of short sequences, %
the constant fudges in all results and the $U$-dependence, %
$M$'s incomputability and crude but practical approximations. %
In short, universal prediction solves or avoids or meliorates many
foundational and philosophical problems, but has to be compromised
in practice.

%%%%%%%%%%%%%%%%%%%%%%%%%%%%%%%%%%%%%%%%%%%%%%%%%%%%%%%%%%%%%%%
%         Bibliography        %
%%%%%%%%%%%%%%%%%%%%%%%%%%%%%%%%%%%%%%%%%%%%%%%%%%%%%%%%%%%%%%%

\begin{small}

\end{small}

\end{document}